\documentclass[runningheads]{llncs}
\usepackage[T1]{fontenc}
\usepackage{graphicx}
\usepackage[misc,geometry]{ifsym}
\usepackage{amsmath,amssymb,amsfonts}
\usepackage{algorithmic}
\usepackage{graphicx}
\usepackage{float}
\usepackage{textcomp}
\usepackage[table]{xcolor}
\usepackage{booktabs}
\usepackage{hyperref}
\usepackage{multirow}
\usepackage{mathrsfs}
\usepackage{url}
\usepackage[perpage]{footmisc}

\usepackage[most]{tcolorbox}
\tcbset{
  mypromptbox/.style={
    colback=blue!5!white,    
    colframe=blue!40!black,  
    boxrule=0.5pt,           
    arc=4pt,                 
    left=6pt,
    right=6pt,
    top=6pt,
    bottom=6pt
  }
}

\usepackage{xcolor}

\begin{document}
\title{Revolutionizing Precise Low Back Pain Diagnosis via Contrastive Learning}


%
%

\author{Thanh Binh Le\inst{1\star} \and
Hoang Nhat Khang Vo \orcidID{0009-0009-7631-2450} \inst{1\star}, Tan Ha Mai \orcidID{0000-0003-4429-2343} \inst{2} \and
(\Letter) Trong Nhan Phan \inst{1}}
\authorrunning{Binh T. Le, Khang H. N. Vo, et al.}
\renewcommand{\thefootnote}{\fnsymbol{footnote}}
\footnotetext[1]{The two authors contributed equally to this paper. The order of authors follows the alphabetical arrangement of their names.}

\institute{Faculty of Computer Science and Engineering, Ho Chi Minh City University of Technology, Vietnam National University (HCMUT-VNU) \\
\email{\{binh.lethanh, khang.vo872003, nhanpt\}@hcmut.edu.vn} \\
\and National Taiwan University \\
\email{maitanha.ai@gmail.com} \\
}

\maketitle              
\begin{abstract}
Low back pain affects millions worldwide, driving the need for robust diagnostic models that can jointly analyze complex medical images and accompanying text reports. We present $\mathtt{LumbarCLIP}$, a novel multimodal framework that leverages contrastive language-image pretraining to align lumbar spine MRI scans with corresponding radiological descriptions. Built upon a curated dataset containing axial MRI views paired with expert-written reports, $\mathtt{LumbarCLIP}$ integrates vision encoders (ResNet-$50$, Vision Transformer, Swin Transformer) with a BERT-based text encoder to extract dense representations. These are projected into a shared embedding space via learnable projection heads - configurable as linear or non-linear—and normalized to facilitate stable contrastive training using a soft CLIP loss. Our model achieves state-of-the-art performance on downstream classification, reaching up to $95.00\%$ accuracy and $94.75\%$ F$1$-score on the test set, despite inherent class imbalance. Extensive ablation studies demonstrate that linear projection heads yield more effective cross-modal alignment than non-linear variants. $\mathtt{LumbarCLIP}$ offers a promising foundation for automated musculoskeletal diagnosis and clinical decision support.

\end{abstract}
\section{Introduction}



Low back pain (LBP) is a leading global health challenge, affecting approximately $628.8$ million people in $2021$, with projections estimating $843$ million cases by $2050$ \cite{Cheng2025}. As the primary cause of disability worldwide, LBP significantly impairs quality of life and imposes substantial burdens on healthcare systems \cite{Cheng2025,DAntoni2022}. Its complex etiology, encompassing spinal abnormalities, visceral diseases, or malignancies, often results in unidentified causes despite advanced diagnostics \cite{Flynn2011ImagingLBP}. This complexity underscores the urgent need for enhanced diagnostic tools to improve accuracy and patient outcomes.

Diagnosing LBP requires integrating patient demographics, physical examinations, imaging studies, and, in some cases, laboratory tests or specialist consultations \cite{Flynn2011ImagingLBP}. \emph{Magnetic Resonance Imaging} (MRI) remains the gold standard for evaluating spinal structures in LBP \cite{DAntoni2022}. However, MRI interpretation is expertise-intensive and susceptible to subjective bias, leading to diagnostic inconsistencies and variable treatment outcomes \cite{Flynn2011ImagingLBP}. Moreover, inappropriate imaging can exacerbate patient outcomes by increasing fear-avoidance behaviors or unnecessary interventions \cite{Flynn2011ImagingLBP}. Consequently, there is a critical need for computer-aided diagnostic systems to support clinicians in achieving objective, consistent, and accurate LBP diagnoses \cite{DAntoni2022}.

To address this, we propose $\mathtt{LumbarCLIP}$, an innovative  way of diagnosing LBP that integrates MRI images and associated text reports. Inspired by multimodal approaches in medical imaging \cite{Acosta2022}, LumbarCLIP leverages advanced image backbones to extract robust visual features. Additionally, we utilize a projection head experimented with different dimensions to enhance the model's performance, building on established techniques in contrastive learning \cite{ZHAO2025103551}. To sum up, our contributions are as follows.
\begin{itemize}
    \item We propose $\mathtt{LumbarCLIP}$ as an alternative approach for LBP detection using multimodal MRI data. An additional projection head was leveraged to improve $\mathtt{LumbarCLIP}$'s diagnostic performance.
    \item We utilized data augmentation to generate synthetic text reports, enhancing the robustness and generalizability of $\mathtt{LumbarCLIP}$ in the presence of limited medical data.
    \item We conduct experiments to evaluate the impact of projection head dimensions and image backbones (including ResNet-$50$, Vision Transformer (ViT), and Swin Transformer) on diagnostic accuracy.
\end{itemize}

The remainder of this paper is organized as follows. Section~\ref{sec:related_work} reviews related works on multimodal LLMs and data augmentation in medical imaging. Section~\ref{sec:methodology} presents the proposed methodology for $\mathtt{LumbarCLIP}$, detailing its architecture and training process. Section~\ref{sec:experiments} describes the experimental setup and results, evaluating the model’s performance. Section~\ref{sec:ablation} provides ablation studies to analyze the impact of key components, such as projection head dimensions and image backbones. Finally, Section~\ref{sec:conclusion} concludes the paper and discusses future research directions.

\section{Related Works}
\label{sec:related_work}

\subsection{AI in Low Back Pain Diagnosis}
Recent advances in artificial intelligence (AI) have shown promise in LBP diagnosis, particularly using MRI. D’Antoni et al. \cite{DAntoni2022} review computer-aided diagnosis (CAD) systems for chronic LBP, covering classification and regression tasks that improve diagnostic precision. Similarly, studies like \cite{Liawrungrueang2025} explore AI methodologies for lumbar degenerative disc disease, a common LBP cause, using deep learning models on MRI data. Other works, such as \cite{Alantari2025}, focus on AI for predicting lumbar spinal stenosis, demonstrating improved diagnostic outcomes. However, these approaches often rely on unimodal data, limiting their ability to integrate diverse clinical information, a gap our multimodal $\mathtt{LumbarCLIP}$ addresses.

\subsection{Multimodal Learning and Contrastive Learning in Medical Domain}
Multimodal AI, combining data sources like images and text, has gained traction in medical diagnostics. Acosta et al. \cite{Acosta2022} discuss multimodal biomedical AI, highlighting its potential to enhance diagnostic accuracy by integrating complementary data. Huang et al. \cite{Huang2024} review the synergy of multimodal data and AI, reporting a $6.4\%$ mean improvement in AUC over unimodal models. In musculoskeletal imaging, Burns et al. \cite{Burns2020} emphasize AI’s transformative potential, particularly for conditions like LBP. 

Contrastive Language-Image Pretraining (CLIP) \cite{DBLP:conf/icml/RadfordKHRGASAM21} models are increasingly applied in medical imaging due to their generalizability and zero-shot capabilities \cite{ZHAO2025103551}. Zhao et al. \cite{ZHAO2025103551} survey CLIP applications, including thoracic disease diagnosis and segmentation, demonstrating robustness with limited labeled data. Works like \cite{Wang2022} adapt CLIP for unpaired medical images and text, while \cite{Zhang2023} use CLIP for organ segmentation and tumor detection. These studies highlight CLIP’s potential but rarely address LBP-specific challenges or explore advanced backbones like ViT and Swin Transformer, which is one of the limitations that our proposed approach aims to overcome. 

\subsection{Projection Heads in Contrastive Learning}
Projection heads are widely used in contrastive learning to map high-dimensional features to a lower-dimensional space, enhancing model performance in tasks like image classification and multimodal learning \cite{10.5555/3524938.3525087}. Chen et al. \cite{10.5555/3524938.3525087} introduce projection heads in SimCLR, demonstrating improved representation learning for visual tasks. In medical imaging, projection heads have been applied to optimize feature alignment in CLIP-based models \cite{ZHAO2025103551}.

\subsection{Data Augmentation by Multimodal Large Language Models}

Recent advancements in multimodal large language models (MLLMs) have enabled synthetic data generation from text and images, addressing data scarcity and privacy concerns in medical AI. Hsieh et al. \cite{HSIEH2025110022} introduce DALL-M for context-aware clinical data augmentation, enhancing diagnostic outcomes, though its reliance on large datasets limits use in low-resource settings. Similarly, Deng et al. \cite{DENG2024103001} develop OphGLM for ophthalmology, which improves diagnostic accuracy but requires further validation for generalizability. Moreover, Thawkar et al. \cite{thawakar-etal-2024-xraygpt} propose XrayGPT for chest radiograph summarization, where synthetic data boosts performance, although risks of bias in LLM-generated content persist.

\section{Proposed Methodology}
\label{sec:methodology}

\begin{figure}[htb]
    \vspace{-10pt}
    \centering
    \includegraphics[width=0.9\linewidth]{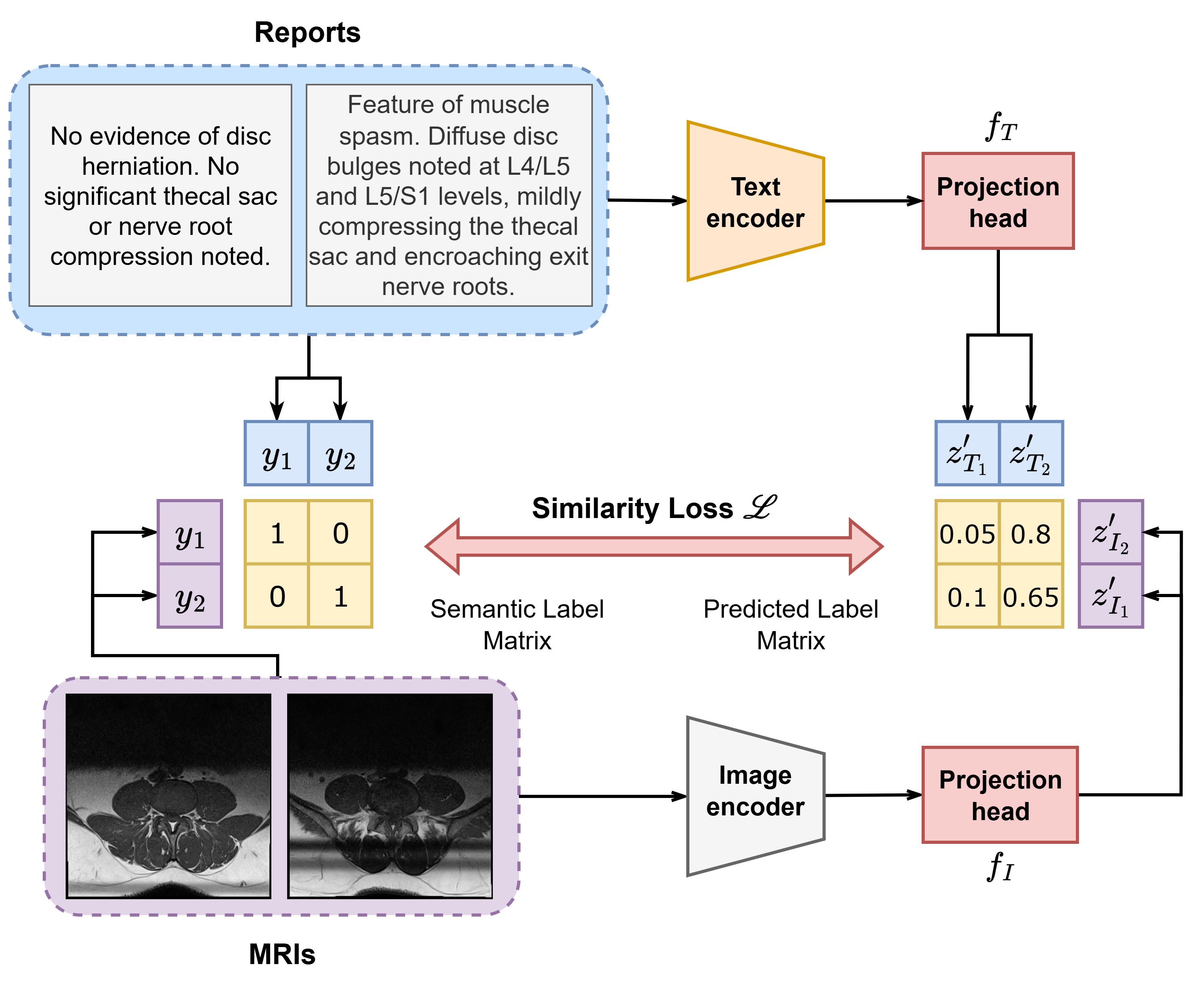}
    \vspace{-5pt}
    \caption{Overview of the \texttt{LumbarCLIP} training pipeline for low back pain diagnosis, which generates synthetic text reports using a multimodal large language model, encodes MRI images into embeddings \(\mathbf{z}_I \in \mathbb{R}^{d_I}\) with vision backbones and texts into \(\mathbf{z}_T \in \mathbb{R}^{d_T}\) with a clinical language model, projects them to a shared latent space as \(\mathbf{z}_I^\prime = f_I(\mathbf{z}_I)\) and \(\mathbf{z}_T^\prime = f_T(\mathbf{z}_T)\) (including augmented texts) via projection heads, and aligns them using a contrastive loss \(\mathcal{L}\) to enhance diagnostic accuracy.}

    \label{fig:pipeline}
    \vspace{-20pt}
\end{figure}

In this study, we introduce \texttt{LumbarCLIP}, a framework that integrates vision and text encoders in a contrastive learning setup. The overall pipeline, illustrated in Figure~\ref{fig:pipeline}, provides a high-level overview of the methodology, encompassing data preparation, preprocessing, model architecture, and training processes. This pipeline leverages paired MRI images and text reports to learn aligned representations in a shared latent space, enabling robust classification of diagnostic classes such as ``LBP'' and ``No Finding''.

\subsection{Data Preparation}

The dataset comprises paired lumbar MRI images and corresponding diagnostic text reports, each annotated with a multi-label binary vector $\mathbf{y} \in \{0, 1\}^2$, representing two diagnostic categories: ``LBP'' and ``No Finding''. Each sample includes:
\begin{itemize}
    \item A lumbar MRI image from axial view.
    \item A radiology report or prompt sentence summarizing the findings.
    \item A binary label vector indicating the presence or absence of each diagnostic class.
\end{itemize}
All samples have complete and verified label vectors, ensuring no missing or ambiguous annotations. To improve textual diversity and enhance the language model’s robustness in interpreting imaging data, we leveraged a state-of-the-art multimodal large language model (MLLM) \textbf{LLaVA-Med v1.5 Mistral-7B} \footnote{\url{microsoft/llava-med-v1.5-mistral-7b}} which integrates both vision and language modalities. This model allows for automatic generation of clinically accurate diagnostic descriptions grounded in visual input. Inference was performed using deterministic decoding (\texttt{temperature = 0.2}, \texttt{num\_beams = 1}, \texttt{max\_new\_tokens = 1024}) to ensure clinical accuracy. The input pipeline utilizes pretrained components from the MLLM  checkpoint, including tokenizer, vision encoder, and language decoder.

Given an MRI image and its original textual caption, we prompted the MLLM to generate a total of five distinct yet semantically consistent sentences. One sentence reproduced the original caption verbatim, while the remaining four offer rephrasings or detailed expansions. The prompt used is defined as:

\begin{tcolorbox}[mypromptbox]
\texttt{Generate 5 distinct, precise, and medically accurate sentences to describe the provided lumbar MRI image. One of the sentences must be the original caption: ``\texttt{\{original\_caption\}}''. The other four sentences should rephrase or expand on the original caption, maintaining its key details (e.g., vertebrae count, disc conditions, Pfirrmann grades, narrowing, bulging, endplate changes) while varying the wording and structure. Ensure all sentences are concise and focus solely on the MRI findings. \textbackslash n\textless image\textgreater}
\end{tcolorbox}

\begin{figure}[ht]
    \centering
    \includegraphics[width=\linewidth]{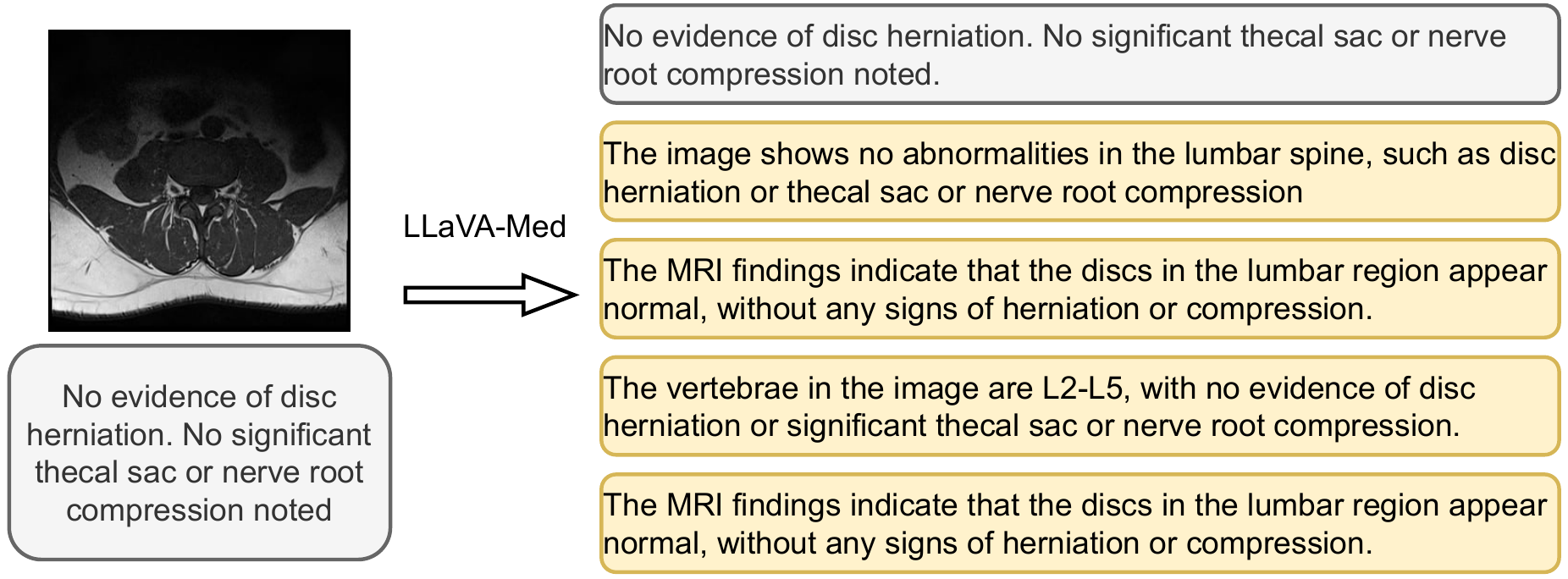}
    \vspace{-8pt}
    \caption{Illustration of our augmentation strategy for each record. We prompted the MLLM to create $4$ new sentences having consistent meaning with the original caption.}
    \label{fig:example}
    \vspace{-10pt}
\end{figure}

Figure \ref{fig:example} illustrates a sample outcome from our augmentation process, showcasing how our MLLM generates synthetic text reports paired with a given MRI image. Following the augmentation process discussed above, the training set expanded from $2,328$ to $11,640$ samples, while the validation set increased from $504$ to $2,520$ samples. This corresponds to a consistent $5$-times increase across both splits.

\subsection{Data Preprocessing}

MRI images are preprocessed to ensure compatibility with deep learning models. They were padded to a square shape, resized to a fixed resolution and normalized. Single-channel images were repeated to form three-channel inputs, aligning with standard vision backbones. Text reports were tokenized using pretrained $\mathtt{BioClinicalBERT}$ tokenizer \cite{alsentzer-etal-2019-publicly}. To enhance robustness, we applied Easy Data Augmentation (EDA) \cite{wei-zou-2019-eda} techniques - synonym replacement, random swap, or random deletion.

\subsection{Training Architecture}

From Figure \ref{fig:pipeline},  
$\mathtt{LumbarCLIP}$ training pipeline comprises three core components:
\begin{itemize}
    \item \textbf{Image encoder}: A backbone (ResNet-$50$, ViT, or Swin Transformer \cite{liu2021Swin}) encodes MRI images into dense embeddings $\mathbf{z}_I \in \mathbb{R}^{d_I}$, where $d_I$ denotes the dimensionality of the image representation space.

    \item \textbf{Text encoder}: We used pretrained $\mathtt{BioClinicalBERT}$ to encode tokenized text reports into dense embeddings $\mathbf{z}_T \in \mathbb{R}^{d_T}$, where $d_T$ denotes the dimensionality of the textual representation space.

    \item \textbf{Projection heads}: Learnable functions project image and text embeddings into a shared latent space of dimension $d$. For an image embedding, the projection head computes
\begin{equation}
    \mathbf{z}_I^\prime = f_I(\mathbf{z}_I),
\end{equation}
    where $f_I: \mathbb{R}^{d_I} \to \mathbb{R}^d$ is either a linear layer or a non-linear sequence consisting of a linear layer, ReLU activation, and another linear layer, followed by $L_2$ normalization. Similarly, for a text embedding
\begin{equation}
    \mathbf{z}_T^\prime = f_T(\mathbf{z}_T),
\end{equation}
    where $f_T: \mathbb{R}^{d_T} \to \mathbb{R}^d$ follows the same configuration. These projection heads, inspired by contrastive learning frameworks, stabilize training by mapping high-dimensional features to a normalized lower-dimensional space, mitigating optimization challenges and enhancing convergence \cite{10.5555/3524938.3525087}. The non-linear configuration introduces additional expressivity, allowing the model to capture complex feature relationships during pretraining. For downstream tasks like classification, the projection heads are discarded, and the encoder outputs $\mathbf{z}_I$ and $\mathbf{z}_T$ are used directly.
\end{itemize}

\subsection{Training Process and Objective Function}
For a batch of $N$ image-text pairs, the model computes embeddings $\{\mathbf{z}_{I_i}^\prime\}_{i=1}^N$ for images, $\{\mathbf{z}_{T_i}^\prime\}_{i=1}^N$ for original texts, and $\{\mathbf{z}_{T_i^\text{aug}}^\prime\}_{i=1}^N$ for augmented texts. Cosine similarities form a matrix $\mathbf{S} \in \mathbb{R}^{N \times N}$, where
\begin{equation}
S_{ij} = \mathbf{z}_{I_i}^\prime \cdot \mathbf{z}_{T_j}^\prime
\end{equation}
A separate similarity matrix is computed for augmented texts.

A label similarity matrix $\mathbf{L} \in \mathbb{R}^{N \times N}$ is defined, where $L_{ij} = \mathbf{y}_i \cdot \mathbf{y}_j$ is the dot product of label vectors for image $i$ and text $j$. This is normalized via softmax to create soft targets
\begin{equation}
    \mathbf{L}^\text{soft}_{ij} = \frac{\exp(L_{ij} / \tau)}{\sum_{k=1}^N \exp(L_{ik} / \tau)},
\end{equation}
where $\tau$ is a temperature parameter. The soft cross-entropy loss is computed bidirectionally:
\begin{itemize}
    \item \textbf{Image-to-Text (I2T) loss} is computed as
\begin{equation}
    \mathcal{L}_\text{I2T} = -\frac{1}{N} \sum_{i=1}^N \sum_{j=1}^N L^\text{soft}_{ij} \log \left( \frac{\exp(S_{ij} / \tau)}{\sum_{k=1}^N \exp(S_{ik} / \tau)} \right)
\end{equation}
    \item \textbf{Text-to-Image (T2I) loss} is computed as
\begin{equation}
    \mathcal{L}_\text{T2I} = -\frac{1}{N} \sum_{i=1}^N \sum_{j=1}^N L^\text{soft}_{ji} \log \left( \frac{\exp(S_{ji} / \tau)}{\sum_{k=1}^N \exp(S_{jk} / \tau)} \right)
\end{equation}
\end{itemize}
An analogous loss \(\mathcal{L}_\text{aug}\) is computed for augmented texts. The total loss $\mathscr{L}$ is defined as
\begin{equation}
\mathscr{L} = \alpha (\mathcal{L}_\text{I2T} + \mathcal{L}_\text{T2I}) + (1 - \alpha) (\mathcal{L}_\text{aug,I2T} + \mathcal{L}_\text{aug,T2I}),
\end{equation}
where \(\alpha \in [0, 1]\) balances the contributions of original and augmented data losses. In our experiments, we set \(\alpha = 0.5\), with the rationale for this choice discussed in Section~\ref{sec:experiments}.

This training process aligns image and text embeddings for both original and augmented data, enabling \(\mathtt{LumbarCLIP}\) to learn robust multimodal representations for low back pain diagnosis. The objective function, through its bidirectional loss components weighted by \(\alpha\), ensures the model effectively captures relationships between MRI images and corresponding text reports, enhancing diagnostic accuracy.

\section{Experiments}
\label{sec:experiments}

\subsection{Dataset Preparation \& Training Configuration}

The dataset, sourced from Mendeley Data \cite{sudirman2019lumbar}, comprises $3,090$ lumbar spine MRI images ($1,545$ T1-weighted and $1,545$ T2-weighted) paired with radiological text reports. After excluding $15$ incomplete or empty reports, $3,360$ valid image-report pairs ($1,680$ T1-weighted and $1,680$ T2-weighted) were retained. Each report is labeled with a binary vector for two classes: ``LBP'' or ``No Finding'', resulting in $2,868$ ``LBP'' and $492$ ``No Finding'' records.

Next, the dataset was split into training ($70\%$), validation ($15\%$), and test ($15\%$) sets with stratified label distribution. By using the MLLM for augmenting our training and validation sets, we expanded the training set to $11,760$ pairs and the validation set to $2,520$ pairs. We also performed upsampling to address imbalance in the augmented dataset. The test set remained unaugmented for unbiased evaluation. Table~\ref{tab:mendeley_split} summarizes the dataset splits.

\begin{table}[t]
\centering
\caption{Distribution of image-report pairs across dataset splits, after augmentation.}
\label{tab:mendeley_split}
\begin{tabular}{p{2.2cm} >{\raggedleft\arraybackslash}p{1.8cm} >{\raggedleft\arraybackslash}p{2.2cm} >{\raggedleft\arraybackslash}p{1.5cm}}
\toprule
\textbf{Dataset Split} & \textbf{Total} & \textbf{No Finding} & \textbf{LBP} \\
\midrule
Training & $11{,}760$ & $1{,}890$ & $9{,}870$ \\
Validation & $2{,}520$ & $405$ & $2{,}115$ \\
Test & $528$ & $66$ & $462$ \\
\midrule
Total & $14{,}808$ & $2{,}361$ & $12{,}447$ \\
\bottomrule
\end{tabular}
\end{table}

Pretraining phase ran for $50$ epochs with a batch size of $128$, using the AdamW optimizer \cite{DBLP:conf/iclr/LoshchilovH19} with a learning rate of $2 \times 10^{-5}$ and weight decay of $1 \times 10^{-4}$. A warmup period over the first $10\%$ of training steps stabilized optimization. The model was trained with the soft CLIP loss, where the parameter $\alpha = 0.5$ equally weights the contributions of original and augmented data losses. This choice ensures that $\mathtt{LumbarCLIP}$ leverages the diversity of augmented text reports while maintaining the fidelity of original data for low back pain diagnosis.

\subsection{Multimodal Classification as Downstream Task}

The downstream task evaluates $\mathtt{LumbarCLIP}$’s pretrained representations through classification of MRI images and text reports into ``LBP'' and ``No Finding'' classes, using a multi-layer perceptron (MLP) classifier. The vision encoder, with projection heads discarded, was frozen during this evaluation to leverage the pretrained embeddings without further adaptation. The MLP processed image embeddings using a single linear layer without activation, directly projecting the embedding dimension to two output units representing the binary classes. The MLP was trained for $50$ epochs with a learning rate of $1 \times 10^{-4}$. 

We evaluated four models from the BenchX framework - ConVIRT \cite{pmlr-v182-zhang22a}, GLoRIA \cite{gloria}, MedCLIP-ViT \cite{Wang2022}, and MedCLIP-ResNet-$50$ \cite{Wang2022} - using an MLP classifier for downstream classification. These models, along with their respective MLP classifiers, were trained on the non-augmented dataset. In contrast, both our proposed \texttt{LumbarCLIP} models and their associated MLP classifiers were trained on the augmented dataset to fully exploit the enriched contrastive supervision. Importantly, the test set remained the same for all models and is not augmented, ensuring fair comparison. To assess the effectiveness of different encoder backbones and projection strategies, we evaluated three variants of \texttt{LumbarCLIP}, each using a vision encoder (ResNet-$50$, ViT, Swin) and linear projection heads of dimension $d=256$.

Table~\ref{tab:benchx_vs_lumbarclip} presents the downstream classification performance of our proposed \texttt{LumbarCLIP} pipeline on the augmented lumbar spine MRI dataset\footnote{All experimental results were obtained by rerunning the models in this study on the augmented dataset.}. All three \texttt{LumbarCLIP} variants achieved strong results, consistently surpassing existing methods. LumbarCLIP-ResNet-$50$-$256$ obtained the highest overall performance, with an accuracy of $95.00\%$ and an F1-score of $94.75\%$. LumbarCLIP-ViT-$256$ followed closely with $94.00\%$ for both metrics, while LumbarCLIP-Swin-$256$ achieved $89.00\%$ accuracy and F$1$-score.

\begin{table*}[t]
\centering
\caption{Comparison of downstream classification performance on the augmented lumbar spine MRI dataset between BenchX models and our \texttt{LumbarCLIP} pipeline.}
\begin{tabular}{lcccc}
\toprule
\textbf{Model} & \textbf{Accuracy} & \textbf{Precision} & \textbf{Recall} & \textbf{F1} \\
\midrule
ConVIRT \cite{pmlr-v182-zhang22a} & $81.52$ & $85.10$ & $94.47$ & $89.54$ \\
GLoRIA \cite{gloria} & $82.25$ & $84.54$ & $96.44$ & $90.09$ \\
MedCLIP-ViT \cite{Wang2022} & $82.94$ & $84.16$ & $98.08$ & $90.59$ \\
MedCLIP-ResNet-$50$ \cite{Wang2022} & $84.24$ & $86.94$ & $95.52$ & $91.03$ \\
\textbf{LumbarCLIP-ResNet50-256 (Ours)} & \cellcolor{blue!10}$\mathbf{95.00}$ & \cellcolor{blue!10}$\mathbf{94.50}$ & \cellcolor{blue!10}$\mathbf{98.08}$ & \cellcolor{blue!10}$\mathbf{94.75}$ \\
\textbf{LumbarCLIP-ViT-256 (Ours)} & $94.00$ & $94.00$ & $94.00$ & $94.00$ \\
\textbf{LumbarCLIP-Swin-256 (Ours)} & $89.00$ & $89.00$ & $90.00$ & $89.00$ \\
\bottomrule
\end{tabular}
\label{tab:benchx_vs_lumbarclip}
\end{table*}

\section{Ablation Studies}
\label{sec:ablation}

\begin{figure}[t]
\vspace{-10pt}
\centering
\includegraphics[width=0.8\linewidth]{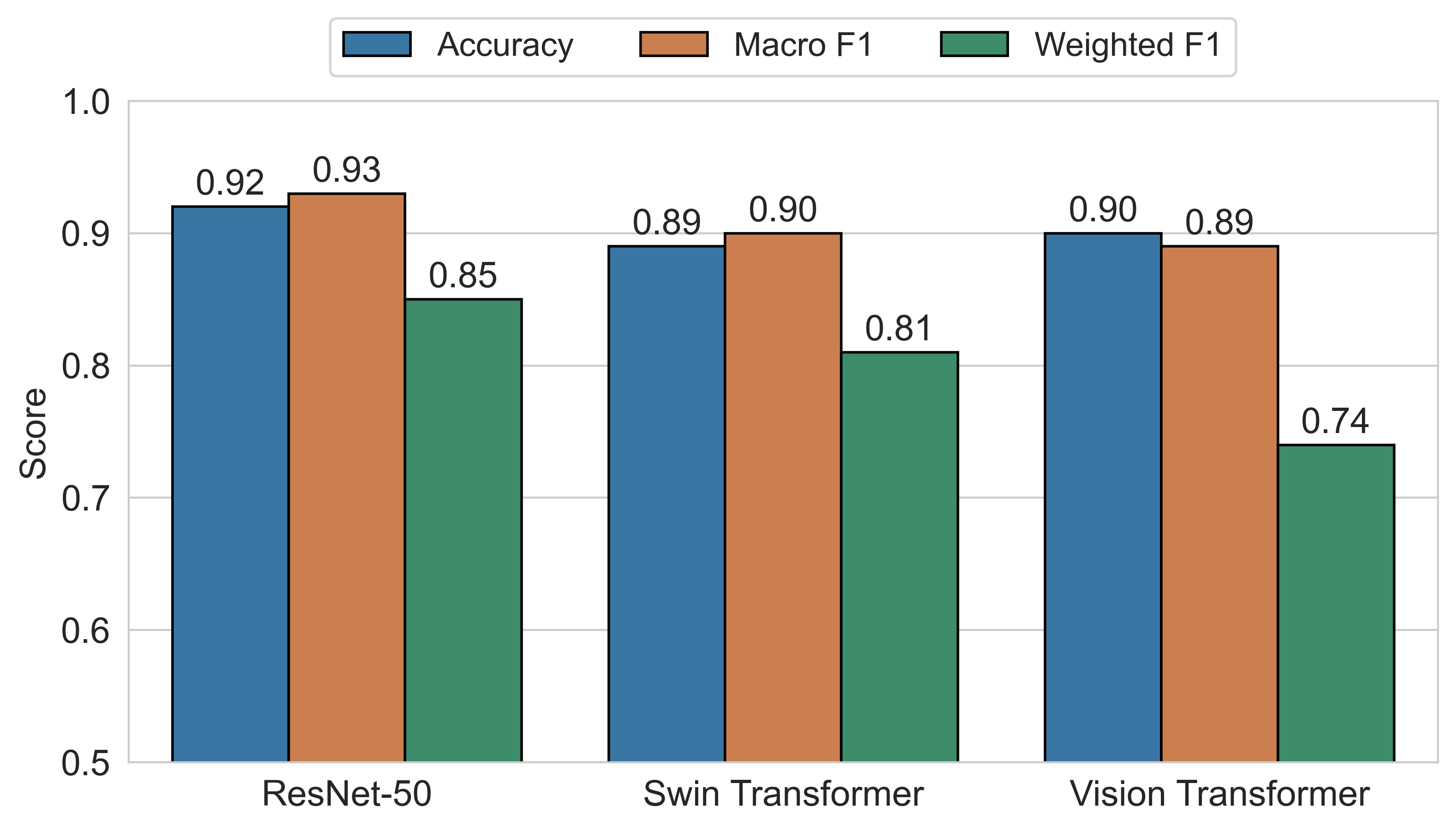}
\vspace{-5pt}
\caption{Classification performance of $\mathtt{LumbarCLIP}$ without projection heads.}
\label{fig:ablation_projection_mendeley}
\vspace{-10pt}
\end{figure}

To assess the role of projection heads in $\mathtt{LumbarCLIP}$’s performance, we conducted ablation studies comparing configurations with and without projection heads during pretraining. Configurations without projection heads followed the original CLIP approach \cite{DBLP:conf/icml/RadfordKHRGASAM21}, using encoder outputs directly for the soft CLIP loss without additional projection layers. When used, projection heads had output dimensions of $256$, $512$, or $1024$ and were either linear or non-linear, improving the matching of image and text data. We tested three vision encoders-ResNet-$50$, ViT, and Swin Transformer-paired with $\mathtt{BioClinicalBERT}$, evaluating performance with an MLP classifier on the augmented Mendeley dataset test set.

Figure~\ref{fig:ablation_projection_mendeley} shows the classification performance of $\mathtt{LumbarCLIP}$ without projection heads. ResNet-$50$ achieves the highest performance with $92\%$ accuracy and a $0.93$ macro F$1$-score, followed by ViT with $90\%$ accuracy and $0.89$ macro F$1$-score, and Swin Transformer with $89\%$ accuracy and $0.90$ macro F1-score. These results indicate solid baseline performance, though Swin Transformer struggles more with minority class detection, likely due to class imbalance in the dataset.

\begin{figure}[t]
\vspace{-10pt}
\centering
\includegraphics[width=\linewidth]{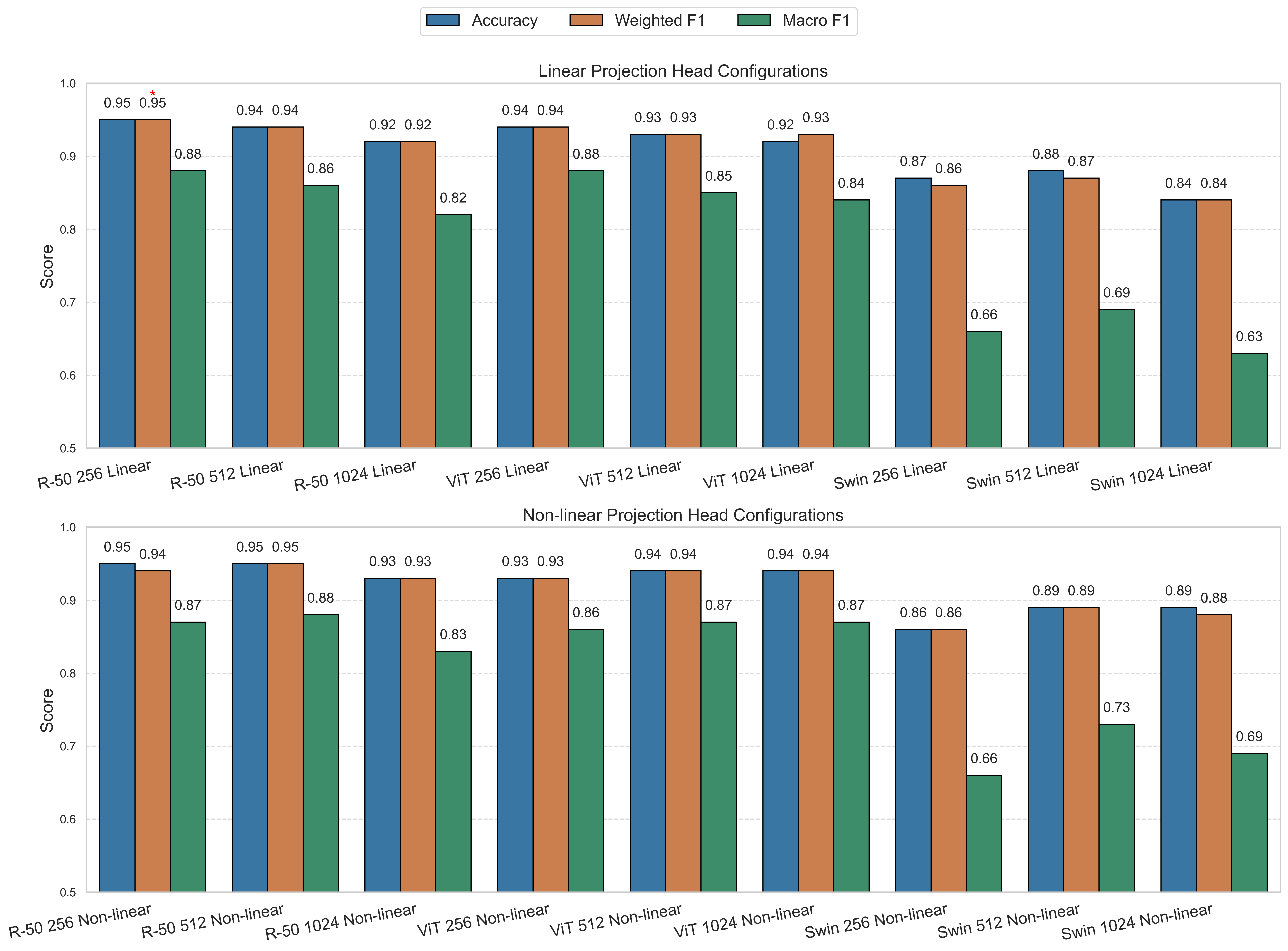}
\caption{Downstream classification performance for $\mathtt{LumbarCLIP}$ across vision encoders and projection head configurations. \textcolor{blue}{Top}: Linear heads. \textcolor{blue}{Bottom}: Non-linear heads.}
\label{fig:result_mendeley}
\vspace{-10pt}
\end{figure}

Figure~\ref{fig:result_mendeley} presents results with projection heads, showing improved performance compared to Figure~\ref{fig:ablation_projection_mendeley}. ResNet-$50$ outperforms ViT and Swin Transformer, achieving up to $95\%$ accuracy and a $0.88$ macro F$1$-score with $256$- and $512$-dimensional heads. Smaller projection heads ($256$, $512$) generally outperform $1024$-dimensional ones, suggesting better matching in lower-dimensional spaces. Linear and non-linear heads perform similarly for ResNet-$50$ and ViT, but non-linear heads slightly boost Swin Transformer’s macro F$1$-score (e.g., $0.73$ vs. $0.69$ at 512 dimensions). Swin Transformer’s macro F1-score ($0.63$–$0.73$) remains lower, likely due to class imbalance affecting minority class detection. Performance drops at $1024$ dimensions suggest possible overfitting or weaker matching. These results highlight the critical role of projection heads, particularly at $256$ and $512$ dimensions, in boosting classification performance.

\section{Conclusion}
\label{sec:conclusion}

This study presents $\mathtt{LumbarCLIP}$, a CLIP-based multimodal framework for low back pain diagnosis, aligning lumbar spine MRIs and radiological reports via soft contrastive pretraining. Trained on axial MRIs, it achieved strong performance in distinguishing LBP from normal cases. Ablation studies highlight the role of projection heads in improving alignment. Future work will address class imbalance, incorporate more MRI modalities, and generalize to broader musculoskeletal applications.

\section*{Acknowledgment}

This research is funded by Vietnam National University - Ho Chi Minh City (VNU-HCM) under grant number: DS$2025$-$20$-$02$

\bibliographystyle{splncs04}
\bibliography{main.bib}

\begin{thebibliography}{10}
\providecommand{\url}[1]{\texttt{#1}}
\providecommand{\urlprefix}{URL }
\providecommand{\doi}[1]{https://doi.org/#1}

\bibitem{Acosta2022}
Acosta, J.N., Falcone, G.J., Rajpurkar, P., Topol, E.J.: Multimodal biomedical ai. Nature Medicine  \textbf{28}(9),  1773--1784 (2022)

\bibitem{Alantari2025}
Al-antari, M.A., Salem, S., Raza, M., Elbadawy, A.S., Bütün, E., Aydin, A.A., Aydoğan, M., Ertuğrul, B., Talo, M., Gu, Y.H.: Evaluating ai-powered predictive solutions for mri in lumbar spinal stenosis: a systematic review. Artificial Intelligence Review  \textbf{58}(8), ~221 (2025)

\bibitem{alsentzer-etal-2019-publicly}
Alsentzer, E., Murphy, J., Boag, W., Weng, W.H., Jindi, D., Naumann, T., McDermott, M.: Publicly available clinical {BERT} embeddings. In: Proceedings of the 2nd Clinical Natural Language Processing Workshop. Association for Computational Linguistics, Minneapolis, Minnesota, USA (2019)

\bibitem{Burns2020}
Burns, J.E., Yao, J., Summers, R.M.: Artificial intelligence in musculoskeletal imaging: A paradigm shift. Journal of Bone and Mineral Research  \textbf{35}(1),  28--35 (2020)

\bibitem{10.5555/3524938.3525087}
Chen, T., Kornblith, S., Norouzi, M., Hinton, G.: A simple framework for contrastive learning of visual representations. In: Proceedings of the 37th International Conference on Machine Learning. ICML'20, JMLR.org (2020)

\bibitem{Cheng2025}
Cheng, M., Xue, Y., Cui, M., Zeng, X., Yang, C., Ding, F., Xie, L.: Global, regional, and national burden of low back pain: Findings from the global burden of disease study 2021 and projections to 2050. Spine  \textbf{50}(7),  E128--E139 (2025)

\bibitem{DAntoni2022}
D'Antoni, F., Russo, F., Ambrosio, L., Borthakur, A., Shetye, S.S., Risbud, M.V., Dionisi, F., Palombi, L., Fiorillo, L., Denaro, V.: Artificial intelligence and computer aided diagnosis in chronic low back pain: A systematic review. International Journal of Environmental Research and Public Health  \textbf{19}(10), ~5971 (2022)

\bibitem{DENG2024103001}
Deng, Z., Gao, W., Chen, C., Niu, Z., Gong, Z., Zhang, R., Cao, Z., Li, F., Ma, Z., Wei, W., Ma, L.: Ophglm: An ophthalmology large language-and-vision assistant. Artificial Intelligence in Medicine  \textbf{157},  103001 (2024)

\bibitem{Flynn2011ImagingLBP}
Flynn, T.W., Smith, B., Chou, R.: Appropriate use of diagnostic imaging in low back pain: a reminder that unnecessary imaging may do as much harm as good. Journal of Orthopaedic \& Sports Physical Therapy  \textbf{41}(11),  838--846 (2011)

\bibitem{HSIEH2025110022}
Hsieh, C., Moreira, C., Nobre, I.B., Sousa, S.C., Ouyang, C., Brereton, M., Jorge, J., Nascimento, J.C.: Dall-m: Context-aware clinical data augmentation with large language models. Computers in Biology and Medicine  \textbf{190},  110022 (2025)

\bibitem{gloria}
Huang, S.C., Shen, L., Lungren, M.P., Yeung, S.: Gloria: A multimodal global-local representation learning framework for label-efficient medical image recognition. In: Proceedings of the IEEE/CVF International Conference on Computer Vision. pp. 3942--3951 (2021)

\bibitem{Huang2024}
Huang, Y.C., Wang, Y.P., Lin, W.C., Fang, J.F., Liao, W.T., Tu, T.Y., Chang, K.V.: A comprehensive review on synergy of multi-modal data and ai technologies in medical diagnosis. Diagnostics  \textbf{11}(3), ~219 (2024)

\bibitem{Liawrungrueang2025}
Liawrungrueang, W., Others: Artificial intelligence-assisted mri diagnosis in lumbar degenerative disc disease: A systematic review. Global Spine Journal  \textbf{15}(2),  1405--1418 (2025)

\bibitem{liu2021Swin}
Liu, Z., Lin, Y., Cao, Y., Hu, H., Wei, Y., Zhang, Z., Lin, S., Guo, B.: Swin transformer: Hierarchical vision transformer using shifted windows. In: Proceedings of the IEEE/CVF International Conference on Computer Vision (ICCV) (2021)

\bibitem{DBLP:conf/iclr/LoshchilovH19}
Loshchilov, I., Hutter, F.: Decoupled weight decay regularization. In: 7th International Conference on Learning Representations, {ICLR} 2019, New Orleans, LA, USA, May 6-9, 2019 (2019)

\bibitem{DBLP:conf/icml/RadfordKHRGASAM21}
Radford, A., Kim, J.W., Hallacy, C., Ramesh, A., Goh, G., Agarwal, S., Sastry, G., Askell, A., Mishkin, P., Clark, J., Krueger, G., Sutskever, I.: Learning transferable visual models from natural language supervision. In: Proceedings of the 38th International Conference on Machine Learning, {ICML} 2021, 18-24 July 2021, Virtual Event. vol.~139, pp. 8748--8763. {PMLR} (2021)

\bibitem{sudirman2019lumbar}
Sudirman, S., Al~Kafri, A., Natalia, F., Meidia, H., Afriliana, N., Al-Rashdan, W., Bashtawi, M., Al-Jumaily, M.: Lumbar spine mri dataset. Mendeley Data, V2 (2019)

\bibitem{thawakar-etal-2024-xraygpt}
Thawakar, O.C., Shaker, A.M., Mullappilly, S.S., Cholakkal, H., Anwer, R.M., Khan, S., Laaksonen, J., Khan, F.: {X}ray{GPT}: Chest radiographs summarization using large medical vision-language models. In: Proceedings of the 23rd Workshop on Biomedical Natural Language Processing. pp. 440--448. Association for Computational Linguistics, Bangkok, Thailand (2024)

\bibitem{Wang2022}
Wang, C., Zhang, Y.: Medclip: Contrastive learning from unpaired medical images and text. NeurIPS Workshops  (2022)

\bibitem{wei-zou-2019-eda}
Wei, J., Zou, K.: {EDA}: Easy data augmentation techniques for boosting performance on text classification tasks. In: Proceedings of the 2019 Conference on Empirical Methods in Natural Language Processing and the 9th International Joint Conference on Natural Language Processing (EMNLP-IJCNLP). pp. 6382--6388. Association for Computational Linguistics, Hong Kong, China (2019)

\bibitem{pmlr-v182-zhang22a}
Zhang, Y., Jiang, H., Miura, Y., Manning, C.D., Langlotz, C.P.: Contrastive learning of medical visual representations from paired images and text. In: Proceedings of the 7th Machine Learning for Healthcare Conference. pp. 2--25. PMLR (2022)

\bibitem{Zhang2023}
Zhang, Y., Liu, H., Hu, J., Liu, J., Yang, J., Jin, Z.: Clip-driven universal model for organ segmentation and tumor detection. ICCV  (2023)

\bibitem{ZHAO2025103551}
Zhao, Z., Liu, Y., Wu, H., Wang, M., Li, Y., Wang, S., Teng, L., Liu, D., Cui, Z., Wang, Q., Shen, D.: Clip in medical imaging: A survey. Medical Image Analysis  \textbf{102},  103551 (2025)

\end{thebibliography}

\end{document}